\documentclass[runningheads]{llncs}
\usepackage{graphicx}
\usepackage{amsmath,amssymb} % define this before the line numbering.
\usepackage{color}
\usepackage{epigraph,booktabs}
\usepackage[width=122mm,left=12mm,paperwidth=146mm,height=193mm,top=12mm,paperheight=217mm]{geometry}

\renewcommand\vec[1]{\mathbf{#1}}

\newcommand{\xSrc}{\vec{x}}
\newcommand{\ySrc}{y}

\newcommand{\xTar}{\vec{u}}

\begin{document}
% \linenumbers
\pagestyle{headings}
\mainmatter
\def\ECCV16SubNumber{5}  % Insert your submission number here

\title{Deep CORAL: Correlation Alignment for Deep Domain Adaptation}

%\titlerunning{ECCV-16 submission ID \ECCV16SubNumber}
%\authorrunning{ECCV-16 submission ID \ECCV16SubNumber}

%\author{Anonymous ECCV submission}
\institute{Paper ID \ECCV16SubNumber}

\author{Baochen Sun\thanks{bsun@cs.uml.edu} and Kate Saenko\thanks{saenko@bu.edu}}
\institute{University of Massachusetts Lowell, Boston University}

\maketitle

\begin{abstract}

Deep neural networks are able to learn powerful representations from large quantities of labeled input data, however they cannot always generalize well across changes in input distributions.
Domain adaptation algorithms have been proposed to compensate for the degradation in performance due to domain shift. In this paper, we address the case when the target domain is unlabeled, requiring unsupervised adaptation. CORAL\cite{coral} is a ``frustratingly easy'' unsupervised domain adaptation method that aligns the second-order statistics of the source and target distributions with a linear transformation. Here, we extend CORAL to learn a nonlinear transformation that aligns correlations of layer activations in deep neural networks (Deep CORAL). Experiments on standard benchmark datasets show state-of-the-art performance.
\end{abstract}

\section{Introduction}
\label{sec:intro}

Many machine learning algorithms assume that the training and test data are independent and identically distributed (i.i.d.). However, this assumption rarely holds in practice as the data is likely to change over time and space. Even though state-of-the-art Deep Convolutional Neural Network features are invariant to low level cues to some degree~\cite{iclr15,iccv15,virtualdataset}, Donahue et al.~\cite{decaf} showed that they still are susceptible to domain shift. Instead of collecting labelled data and training a new classifier for every possible scenario, unsupervised domain adaptation methods~\cite{saenko2010adapting,gfk,sa,bmvc14,bmvc15,coral} try to compensate for the degradation in performance by transferring knowledge from labelled source domains to unlabelled target domains. A recently proposed CORAL method~\cite{coral} aligns the second-order statistics of the source and target distributions with a linear transformation. Even though it is ``frustratingly easy'', it works well for unsupervised domain adaptation. However, it relies on a linear transformation and is not end-to-end: it needs to first extract features, apply the transformation, and then train an SVM classifier in a separate step.

In this work, we extend CORAL to incorporate it directly into deep networks by constructing a differentiable loss function that minimizes the difference between source and target correlations--the CORAL loss. Compared to CORAL, our proposed Deep CORAL approach learns a non-linear transformation that is more powerful and also works seamlessly with deep CNNs. We evaluate our method on standard benchmark datasets and show state-of-the-art performance.
\section{Related Work}
\label{sec:related}

Previous techniques for unsupervised adaptation consisted of re-weighting the training point losses to more closely reflect those in the test distribution~\cite{jiang-zhai07,huang_nips06} or finding a transformation in a lower-dimensional manifold that brings the source and target subspaces closer together. Re-weighting based approaches often assume a restricted form of domain shift--selection bias--and are thus not applicable to more general scenarios. Geodesic methods~\cite{gopalan-iccv11,gfk} bridge the source and target domain by projecting source and target onto points along a geodesic path~\cite{gopalan-iccv11}, or finding a closed-form linear map that transforms source points to target~\cite{gfk}. ~\cite{outlooks,sa} align the subspaces by computing the linear map that minimizes the Frobenius norm of the difference between the top $n$ eigenvectors. In contrast, CORAL~\cite{coral} minimizes domain shift by aligning the second-order statistics of source and target distributions.

Adaptive deep neural networks have recently been explored for unsupervised adaptation. DLID~\cite{chopra2013dlid} trains a joint source and target CNN architecture with two adaptation layers. 
DDC~\cite{tzeng_arxiv15} applies a single linear kernel to one layer to minimize Maximum Mean Discrepancy (MMD) while DAN~\cite{dan_long15} minimizes MMD with multiple kernels applied to multiple layers. ReverseGrad~\cite{reversegrad} adds a binary classifier to explicitly confuse the two domains.

Our proposed Deep CORAL approach is similar to DDC, DAN, and ReverseGrad in the sense that a new loss (CORAL loss) is added to minimize the difference in learned feature covariances across domains, which is similar to minimizing MMD with a polynomial kernel. However, it is more powerful than DDC (which aligns sample means only), much simpler to optimize than DAN and ReverseGrad, and can be integrated into different layers or architectures seamlessly.

\section{Deep CORAL}
\label{sec:methods}

\begin{figure}
\centering
\includegraphics[width=0.9\linewidth]{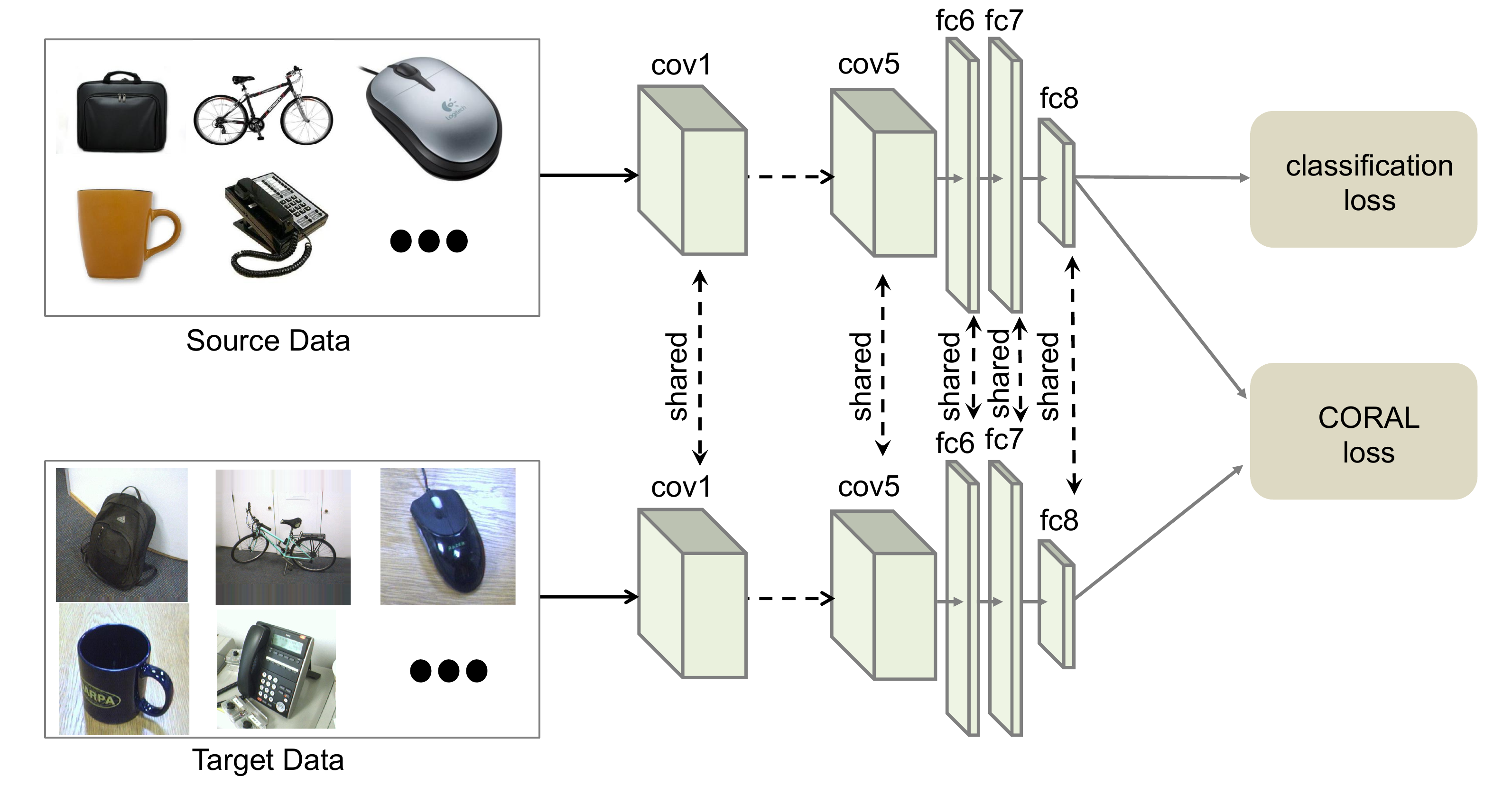}
\caption{\small Sample Deep CORAL architecture based on a CNN with a classifier layer. For generalization and simplicity, here we apply the CORAL loss to the $fc8$ layer of AlexNet~\cite{alexnet}. Integrating it to other layers or network architectures should be straightforward.}
\label{fig:d-coral}
\end{figure}

We address the unsupervised domain adaptation scenario where there are no labelled training data in the target domain, and propose to leverage both the deep features pre-trained on a large generic domain (such as Imagenet~\cite{imagenet}) and the labelled source data. In the meantime, we also want the final learned features to work well on the target domain. The first goal can be achieved by initializing the network parameters from the generic pre-trained network and fine-tuning it on the labelled source data. For the second goal, we propose to minimize the difference in second-order statistics between the source and target feature activations, i.e. the CORAL loss. Figure~\ref{fig:d-coral} shows a sample Deep CORAL architecture using our proposed correlation alignment layer for deep domain adaptation. We refer to Deep CORAL as any deep network incorporating the CORAL loss for domain adaptation. 

\subsection{CORAL Loss}
We first describe the CORAL loss between two domains for a single feature layer.
Suppose we are given source-domain training examples $D_S=\{\xSrc_i\}$, $\xSrc\in\mathbb{R}^d$ with labels $L_S=\{\ySrc_i\}$, $i \in\{1,...,L\}$, and unlabeled target data $D_T=\{\xTar_i\}$, $\xTar \in \mathbb{R}^d$. Suppose the number of source and target data are $n_{S}$ and $n_{T}$ respectively. Here both $\xSrc$ and $\xTar$ are the $d$-dimensional deep layer activations $\phi(I)$ of input $I$ that we are trying to learn. 
Suppose $D_S^{ij}$ ($D_T^{ij}$) indicates the $j$-th dimension of the $i$-th source (target) data example and $C_{S}$ ($C_{T}$) denote the feature covariance matrices. 

We define the CORAL loss as the distance between the second-order statistics (covariances) of the source and target features:
      \begin{equation}
      \begin{aligned}
      {\ell_{CORAL}}= {\frac{1}{4d^2}}{\| C_{S} - C_{T} \|}^2_F\\
      \end{aligned}
      \label{eq:coral}
      \end{equation}
where ${\|\cdot\|}^2_F$ denotes the squared matrix Frobenius norm. 
The covariance matrices of the source and target data are given by:
      \begin{equation}
      \begin{aligned}
     % &C_{S} = {\frac{1}{n_{S}-1}}{\sum_{i=1}^{n_{S}} (\xSrc_i-\mu_s)(\xSrc_i-\mu_s)^{\top}}\\
      &C_{S}= {\frac{1}{n_{S}-1}}({D_S^{\top} D_S - \frac{1}{n_{S}}{({\textbf{1}}^{\top}D_S})^{\top}{({\textbf{1}}^{\top}D_S})})
      \end{aligned}
      \label{eq:cov_s}
      \end{equation}
      \begin{equation}
      \begin{aligned}
      %&C_{T} = {\frac{1}{n_{T}-1}}{\sum_{i=1}^{n_{T}} (\xTar_i-\mu_t)(\xTar_i-\mu_t)^{\top}}\\
      &C_{T}= {\frac{1}{n_{T}-1}}({D_T^{\top} D_T - \frac{1}{n_{T}}{({\textbf{1}}^{\top}D_T})^{\top}{({\textbf{1}}^{\top}D_T})})
      \end{aligned}
      \label{eq:cov_t}
      \end{equation}
where $\textbf{1}$ is a column vector with all elements equal to 1. 

The gradient with respect to the input features can be calculated using the chain rule:
      \begin{equation}
      \begin{aligned}
      &\frac{\partial{\ell_{CORAL}}}{\partial{D_S^{ij}}}=\frac{1}{d^2(n_S-1)}((D_S^{\top}-\frac{1}{n_{S}}({{\textbf{1}}^{\top}D_S})^{\top}{\textbf{1}}^{\top})^{\top}(C_{S} - C_{T}))^{ij}
      \end{aligned}
      \label{eq:gradient_s}
      \end{equation}
      \begin{equation}
      \begin{aligned}
      \frac{\partial{\ell_{CORAL}}}{\partial{D_T^{ij}}}=-\frac{1}{d^2(n_T-1)}((D_T^{\top}-\frac{1}{n_{T}}({{\textbf{1}}^{\top}D_T})^{\top}{\textbf{1}}^{\top})^{\top}(C_{S} - C_{T}))^{ij}
      \end{aligned}
      \label{eq:gradient_t}
      \end{equation}
We use batch covariances and the network parameters are shared between two networks.

\subsection{End-to-end Domain Adaptation with CORAL Loss}
We describe our method by taking a multi-class classification problem as the running example. As mentioned before, the final deep features need to be both discriminative enough to train strong classifier and invariant to the difference between source and target domains. Minimizing the classification loss itself is likely to lead to overfitting to the source domain, causing reduced performance on the target domain. On the other hand, minimizing the CORAL loss alone might lead to degenerated features. For example, the network could project all of the source and target data to a single point, making the CORAL loss trivially zero. However, no strong classifier can be constructed on these features. Joint training with both the classification loss and CORAL loss is likely to learn features that work well on the target domain:
      \begin{equation}
      \begin{aligned}
      {\ell}= {\ell_{CLASS.}} + \sum_{i=1}^{t}\lambda_{i}{\ell_{CORAL}}\\
      \end{aligned}
      \label{eq:obj}
      \end{equation}
where $t$ denotes the number of CORAL loss layers in a deep network and $\lambda$ is a weight that trades off the adaptation with classification accuracy on the source domain. As we show below, these two losses play counterparts and reach an~\emph{equilibrium} at the end of training, where the final features are expected to work well on the target domain.

 \section{Experiments}
\label{sec:exp}

We evaluate our method on a standard domain adaptation benchmark -- the Office dataset~\cite{saenko2010adapting}. The Office dataset contains 31 object categories from an office environment in 3 image domains: $Amazon$, $DSLR$,  and $Webcam$. 

We follow the standard protocol of~\cite{gfk,dan_long15,decaf,tzeng_arxiv15,reversegrad} and use all the labelled source data and all the target data without labels. 
Since there are 3 domains, we conduct experiments on all 6 shifts, taking one domain as the source and another as the target. 

In this experiment, we apply the CORAL loss to the last classification layer
as it is the most general case--most deep classifier architectures (e.g., convolutional, recurrent) contain a fully connected layer for classification. Applying the CORAL loss to other layers or other network architectures should be straightforward. 

The dimension of last fully connected layer ($fc8$) was set to the number of categories (31) and initialized with $\mathcal{N}(0,0.005)$. The learning rate of $fc8$ is set to 10 times the other layers as it was training from scratch. We initialized the other layers with the parameters pre-trained on ImageNet~\cite{imagenet} and kept the original layer-wise parameter settings. In the training phase, we set the batch size to 128, base learning rate to $10^{-3}$, weight decay to $5\times10^{-4}$, and momentum to 0.9. The weight of the CORAL loss ($\lambda$) is set in such way that at the end of training the classification loss and CORAL loss are roughly the same. It seems be a reasonable choice as we want to have a feature representation that is both discriminative and also minimizes the distance between the source and target domains. We used Caffe~\cite{caffe} and BVLC Reference CaffeNet for all of our experiments.

We compare to 7 recently published methods: CNN~\cite{alexnet} (no adaptation), GFK~\cite{gfk}, SA~\cite{sa}, TCA~\cite{tca}, CORAL~\cite{coral}, DDC~\cite{tzeng_arxiv15}, DAN~\cite{dan_long15}. GFK, SA, and TCA are manifold based methods that project the source and target distributions into a lower-dimensional manifold and are not end-to-end deep methods. DDC adds a domain confusion loss to AlexNet and fine-tunes it on both the source and target domain. DAN is similar to DDC but utilizes a multi-kernel selection method for better mean embedding matching and adapts in multiple layers. For direct comparison, DAN in this paper uses the hidden layer $fc8$. For GFK, SA, TCA, and CORAL, we use the $fc7$ feature fine-tuned on the source domain ($FT7$ in~\cite{coral}) as it achieves better performance than generic pre-trained features, and train a linear SVM~\cite{sa,coral}. To have a fair comparison, we use accuracies reported by other authors with exactly the same setting or conduct experiments using the source code provided by the authors.

From Table~\ref{tab:result_office31} we can see that Deep CORAL (D-CORAL) achieves better average performance than CORAL and the other 6 baseline methods. In three 3 out of 6 shifts, it achieves the highest accuracy. For the other 3 shifts, the margin between D-CORAL and the best baseline method is very small ($\leqslant0.7$).

\begin{table}
\begin{center}
\resizebox{0.98\columnwidth}{!}{
\begin{tabular}{|l||c|c|c|c|c|c|c|}
\hline
~ & A$\rightarrow$D&	A$\rightarrow$W	&D$\rightarrow$A&	D$\rightarrow$W	&W$\rightarrow$A	&W$\rightarrow$D&AVG\\ 
\hline
GFK	&52.4\scriptsize{$\pm$0.0}	&54.7\scriptsize{$\pm$0.0}	&43.2\scriptsize{$\pm$0.0}	&92.1\scriptsize{$\pm$0.0}	&41.8\scriptsize{$\pm$0.0}	&96.2\scriptsize{$\pm$0.0}  &63.4\\
\hline
SA	&50.6\scriptsize{$\pm$0.0}	&47.4\scriptsize{$\pm$0.0}	&39.5\scriptsize{$\pm$0.0}	&89.1\scriptsize{$\pm$0.0}	&37.6\scriptsize{$\pm$0.0}	&93.8\scriptsize{$\pm$0.0}  &59.7\\
\hline
TCA	&46.8\scriptsize{$\pm$0.0}	&45.5\scriptsize{$\pm$0.0}	&36.4\scriptsize{$\pm$0.0}	&81.1\scriptsize{$\pm$0.0}	&39.5\scriptsize{$\pm$0.0}	&92.2\scriptsize{$\pm$0.0}  &56.9\\
\hline
CORAL	&65.7\scriptsize{$\pm$0.0}	&64.3\scriptsize{$\pm$0.0}	&48.5\scriptsize{$\pm$0.0}	&\textbf{96.1}\scriptsize{$\pm$0.0}	&48.2\scriptsize{$\pm$0.0}	&\textbf{99.8}\scriptsize{$\pm$0.0}  &70.4\\
\hline
CNN	 &63.8\scriptsize{$\pm$0.5}	&61.6\scriptsize{$\pm$0.5}	&51.1\scriptsize{$\pm$0.6}	&95.4\scriptsize{$\pm$0.3}	&49.8\scriptsize{$\pm$0.4}	&99.0\scriptsize{$\pm$0.2}  &70.1\\
\hline
DDC	 &64.4\scriptsize{$\pm$0.3}	&61.8\scriptsize{$\pm$0.4}	&52.1\scriptsize{$\pm$0.8}	&95.0\scriptsize{$\pm$0.5}	&\textbf{52.2}\scriptsize{$\pm$0.4}	&98.5\scriptsize{$\pm$0.4}  &70.6\\
\hline
DAN	 &65.8\scriptsize{$\pm$0.4}	&63.8\scriptsize{$\pm$0.4}	&\textbf{52.8}\scriptsize{$\pm$0.4}	&94.6\scriptsize{$\pm$0.5}	&51.9\scriptsize{$\pm$0.5}	&98.8\scriptsize{$\pm$0.6}  &71.3\\
\hline
D-CORAL	&\textbf{66.8}\scriptsize{$\pm$0.6}	&\textbf{66.4}\scriptsize{$\pm$0.4}	
&\textbf{52.8}\scriptsize{$\pm$0.2}	
&95.7\scriptsize{$\pm$0.3}	
&51.5\scriptsize{$\pm$0.3}	
&99.2\scriptsize{$\pm$0.1}  &\textbf{72.1}\\
\hline
\end{tabular}
}
\end{center}
\caption{Object recognition accuracies for all 6 domain shifts on the standard Office dataset with deep features, following the standard unsupervised adaptation protocol. }
\label{tab:result_office31}
\vspace{-0.1in}
\end{table}

\begin{figure}
\centering
\includegraphics[width=\linewidth]{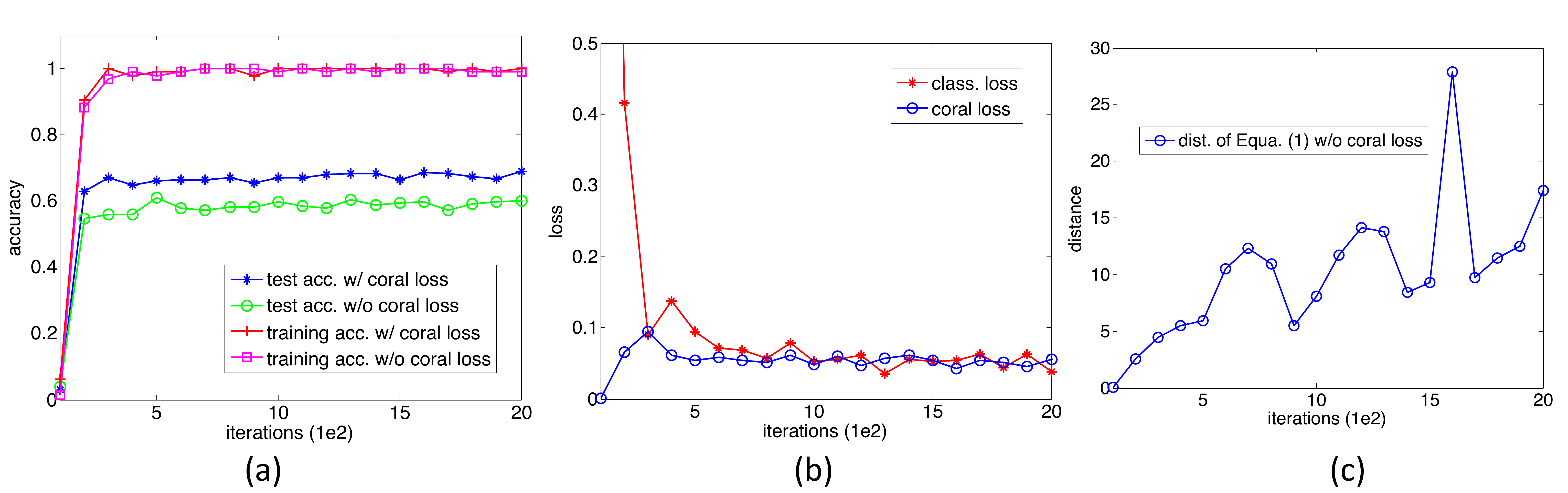}
\caption{Detailed analysis of shift A$\rightarrow$W for training w/ vs. w/o CORAL loss. (a): training and test accuracies for training w/ vs. w/o CORAL loss. We can see that adding CORAL loss helps achieve much better performance on the target domain while maintaining strong classification accuracy on the source domain. (b): classification loss and CORAL loss for training w/ CORAL loss. As the last fully connected layer is randomly initialized with $\mathcal{N}(0,0.005)$, CORAL loss is very small while classification loss is very large at the beginning. After training for a few hundred iterations, these two losses are about the same. (c): CORAL distance for training w/o CORAL loss  (setting the weight to 0). The distance is getting much larger ($\geqslant100$ times larger compared to training w/ CORAL loss).}
\label{fig:a_w}
\end{figure}

To get a better understanding of Deep CORAL, we generate three plots for domain shift A$\rightarrow$W. In Figure~\ref{fig:a_w}(a) we show the training (source) and testing (target) accuracies for training with vs. without CORAL loss. We can clearly see that adding the CORAL loss helps achieve much better performance on the target domain while maintaining strong classification accuracy on the source domain. 

In Figure~\ref{fig:a_w}(b) we visualize both the classification loss and the CORAL loss for training w/ CORAL loss. As the last fully connected layer is randomly initialized with $\mathcal{N}(0,0.005)$,  in the beginning the CORAL loss is very small while the classification loss is very large. After training for a few hundred iterations, these two losses are about the same. In Figure~\ref{fig:a_w}(c) we show the CORAL distance between the domains for training w/o CORAL loss (setting the weight to 0). We can see that the distance is getting much larger ($\geqslant100$ times larger compared to training w/ CORAL loss). Comparing Figure~\ref{fig:a_w}(b) and Figure~\ref{fig:a_w}(c), we can see that even though the CORAL loss is not always decreasing during training, if we set its weight to 0, the distance between source and target domains becomes much larger. This is reasonable as fine-tuning without domain adaptation is likely to overfit the features to the source domain. Our CORAL loss constrains the distance between source and target domain during the fine-tuning process and helps to maintain an~\emph{equilibrium} where the final features work well on the target domain.

\section{Conclusion}
\label{sec:concl}

In this work, we extended CORAL, a simple yet effective unsupervised domain adaptation method, to perform end-to-end adaptation in deep neural networks. Experiments on standard benchmark datasets show state-of-the-art performance. Deep CORAL works seamlessly with deep networks and can be easily integrated into different layers or network architectures.

\clearpage
\bibliographystyle{splncs}
\bibliography{main.bib}

\begin{thebibliography}{10}

\bibitem{coral}
Sun, B., Feng, J., Saenko, K.:
\newblock Return of frustratingly easy domain adaptation.
\newblock In: AAAI. (2016)

\bibitem{iclr15}
Peng, X., Sun, B., Ali, K., Saenko, K.:
\newblock What do deep cnns learn about objects?
\newblock In: ICLR Workshop Track. (2015)

\bibitem{iccv15}
Peng, X., Sun, B., Ali, K., Saenko, K.:
\newblock Learning deep object detectors from 3d models.
\newblock In: ICCV. (2015)

\bibitem{virtualdataset}
Sun, B., Peng, X., Saenko, K.:
\newblock Generating large scale image datasets from 3d cad models.
\newblock In: CVPR'15 Workshop on The Future of Datasets in Vision. (2015)

\bibitem{decaf}
Donahue, J., Jia, Y., Vinyals, O., Hoffman, J., Zhang, N., Tzeng, E., Darrell,
  T.:
\newblock Decaf: A deep convolutional activation feature for generic visual
  recognition.
\newblock In: ICML. (2014)

\bibitem{saenko2010adapting}
Saenko, K., Kulis, B., Fritz, M., Darrell, T.:
\newblock Adapting visual category models to new domains.
\newblock In: ECCV.
\newblock (2010)

\bibitem{gfk}
Gong, B., Shi, Y., Sha, F., Grauman, K.:
\newblock Geodesic flow kernel for unsupervised domain adaptation.
\newblock In: CVPR. (2012)

\bibitem{sa}
Fernando, B., Habrard, A., Sebban, M., Tuytelaars, T.:
\newblock Unsupervised visual domain adaptation using subspace alignment.
\newblock In: ICCV. (2013)

\bibitem{bmvc14}
Sun, B., Saenko, K.:
\newblock From virtual to reality: Fast adaptation of virtual object detectors
  to real domains.
\newblock In: BMVC. (2014)

\bibitem{bmvc15}
Sun, B., Saenko, K.:
\newblock Subspace distribution alignment for unsupervised domain adaptation.
\newblock In: BMVC. (2015)

\bibitem{jiang-zhai07}
Jiang, J., Zhai, C.:
\newblock {Instance Weighting for Domain Adaptation in NLP}.
\newblock In: ACL. (2007)

\bibitem{huang_nips06}
Huang, J., Smola, A.J., Gretton, A., Borgwardt, K.M., Sch{\"o}lkopf, B.:
\newblock Correcting sample selection bias by unlabeled data.
\newblock In: NIPS. (2006)

\bibitem{gopalan-iccv11}
Gopalan, R., Li, R., Chellappa, R.:
\newblock Domain adaptation for object recognition: An unsupervised approach.
\newblock In: ICCV. (2011)

\bibitem{outlooks}
Harel, M., Mannor, S.:
\newblock Learning from multiple outlooks.
\newblock In: ICML. (2011)

\bibitem{chopra2013dlid}
Chopra, S., Balakrishnan, S., Gopalan, R.:
\newblock Dlid: Deep learning for domain adaptation by interpolating between
  domains.
\newblock In: ICML Workshop. (2013)

\bibitem{tzeng_arxiv15}
Tzeng, E., Hoffman, J., Zhang, N., Saenko, K., Darrell, T.:
\newblock Deep domain confusion: Maximizing for domain invariance.
\newblock CoRR \textbf{abs/1412.3474} (2014)

\bibitem{dan_long15}
Long, M., Cao, Y., Wang, J., Jordan, M.I.:
\newblock Learning transferable features with deep adaptation networks.
\newblock In: ICML. (2015)

\bibitem{reversegrad}
Ganin, Y., Lempitsky, V.:
\newblock Unsupervised domain adaptation by backpropagation.
\newblock In: ICML. (2015)

\bibitem{imagenet}
Deng, J., Dong, W., Socher, R., Li, L.J., Li, K., Fei-Fei, L.:
\newblock Imagenet: A large-scale hierarchical image database.
\newblock In: CVPR. (2009)

\bibitem{alexnet}
Krizhevsky, A., Sutskever, I., Hinton, G.E.:
\newblock Imagenet classification with deep convolutional neural networks.
\newblock In: NIPS. (2012)

\bibitem{caffe}
Jia, Y., Shelhamer, E., Donahue, J., Karayev, S., Long, J., Girshick, R.,
  Guadarrama, S., Darrell, T.:
\newblock Caffe: Convolutional architecture for fast feature embedding.
\newblock arXiv preprint arXiv:1408.5093 (2014)

\bibitem{tca}
Pan, S.J., Tsang, I.W., Kwok, J.T., Yang, Q.:
\newblock Domain adaptation via transfer component analysis.
\newblock In: IJCAI. (2009)

\end{thebibliography}
\end{document}